# Bioinspired soft robotics: How do we learn from creatures?

Yang Yang, Zhiguo He, Pengcheng Jiao, and Hongliang Ren, *Senior Member, IEEE*

**Abstract**—**Soft robotics has opened a unique path to flexibility and environmental adaptability, learning from nature and reproducing biological behaviors. Nature implies answers for how to apply robots to real life. To find out how we learn from creatures to design and apply soft robots, in this Review, we propose a classification method to summarize soft robots based on different functions of biological systems: self-growing, self-healing, self-responsive, and self-circulatory. The bio-function based classification logic is presented to explain *why* we learn from creatures. State-of-art technologies, characteristics, pros, cons, challenges, and potential applications of these categories are analyzed to illustrate *what* we learned from creatures. By intersecting these categories, the existing and potential bio-inspired applications are overviewed and outlooked to finally find the answer, that is, *how* we learn from creatures.**

This work is submitted on March 2nd 2022. Corresponding author: *Hongliang Ren.*

H. Ren is with the Department of Electronic Engineering, Faculty of Engineering, The Chinese University of Hong Kong, Hong Kong 999077, China; Department of Biomedical Engineering (BME), National University of Singapore, Singapore 119077, Singapore; Research Institute, National University of Singapore (Suzhou), Suzhou 215123, Jiangsu, China; Shun Hing Institute of Advanced Engineering, The Chinese University of Hong Kong (CUHK), Hong Kong 999077, Hong Kong, China. (Email:hlren@ieee.org)

Y. Yang is with the Department of Electronic Engineering, Faculty of Engineering, The Chinese University of Hong Kong, Hong Kong 999077; The Institute of Port, Coastal and Offshore Engineering, Ocean College, Zhejiang University, Zhoushan 316021, Zhejiang, China, and Department of Biomedical Engineering (BME), National University of Singapore, Singapore 119077, Singapore.

Z. He is with the Hainan Institute, Zhejiang University, Sanya 572000, Hainan, China, and the Institute of Port, Coastal and Offshore Engineering, Ocean College, Zhejiang University, Zhoushan 316021, Zhejiang, China.

P. Jiao is with the Institute of Port, Coastal and Offshore Engineering, Ocean College, Zhejiang University, Zhoushan 316021, Zhejiang, China.

## I. INTRODUCTION

SOFT robotics [1]-[2], the soft material science-based discipline, focuses more on performing environment-adaptable tasks than rigid robotics, experiencing rapid development [3]-[4]. Given the growing interests (e.g., in biomedical engineering, field exploring, and mechanical automation engineering [5]-[7]), it is important to revisit the soft robotics classification. Developing environmental adaptive, highly flexible, and strongly nonlinear actuation technologies involve the large deformation theory [8], the control science [9], the mechanics [10], and various types of physic sciences such as magnetism, the explosion mechanics, and optics [11]-[14].

The wide range of disciplines involved in soft robotics, brought diversity in its classification. According to actuation materials, it can be classified as the shape memory actuation (SM-series containing polymer and alloy) [14], the ion-exchange polymer-metal composites actuation [15], the dielectric elastomer actuation [16], the magnetic actuation [17], the wire actuation [18], the optical-sensitive materials actuation [19], and pneumatic actuation methods [20], etc. According to robot tasks, it can be classified as grippers [21], crawling soft robots [22]-[23], jumping soft robots [23]-[24], swimming soft robots [25]-[26], and soft manipulators [27]-[28], etc.

As the guide and inspiration sources, natural organisms are diverse and have several constant bio-functions. Learning from creatures, we can classify existing soft robots by these basic bio-functions, which help us obtain ideas, summarize design concepts, and apply robots in a targeted way. In this Review, focusing on bioinspiration, we propose a new classification of soft robots, which contains four categories: the self-growing, self-healing, self-responsive, and self-circulatory soft robots. We present the logic of the bio-function-based classification to explain why we learn from creatures. We survey their state-of-art technologies, characteristics, pros, cons, challenges, and potential applications to illustrate the reasonability and unity of the bio-function-based classification. Namely, what can we learn from creatures? We finally extend beyond these categories to demonstrate existing applications, potential applications, and future development of soft robots. Namely, how can we learn from creatures?

## II. WHY DO WE LEARN FROM CREATURES?

Just as biologists classify living organisms, roboticists have been trying to establish classification systems for soft robots to understand better, design, and apply them. Existing reviews provide in-depth and detailed guidance by classifying soft robots based on actuations, applications, materials, and manufacturing methods (Table I). However, due to the diversity and complexity of the soft robotics research, a bio-function-based classification method has not been proposed yet, with which we can further foresee futural developments of soft robots, such as: how to design soft robots to interact with the environment more reasonably, and how to apply them to more practical application scenarios.

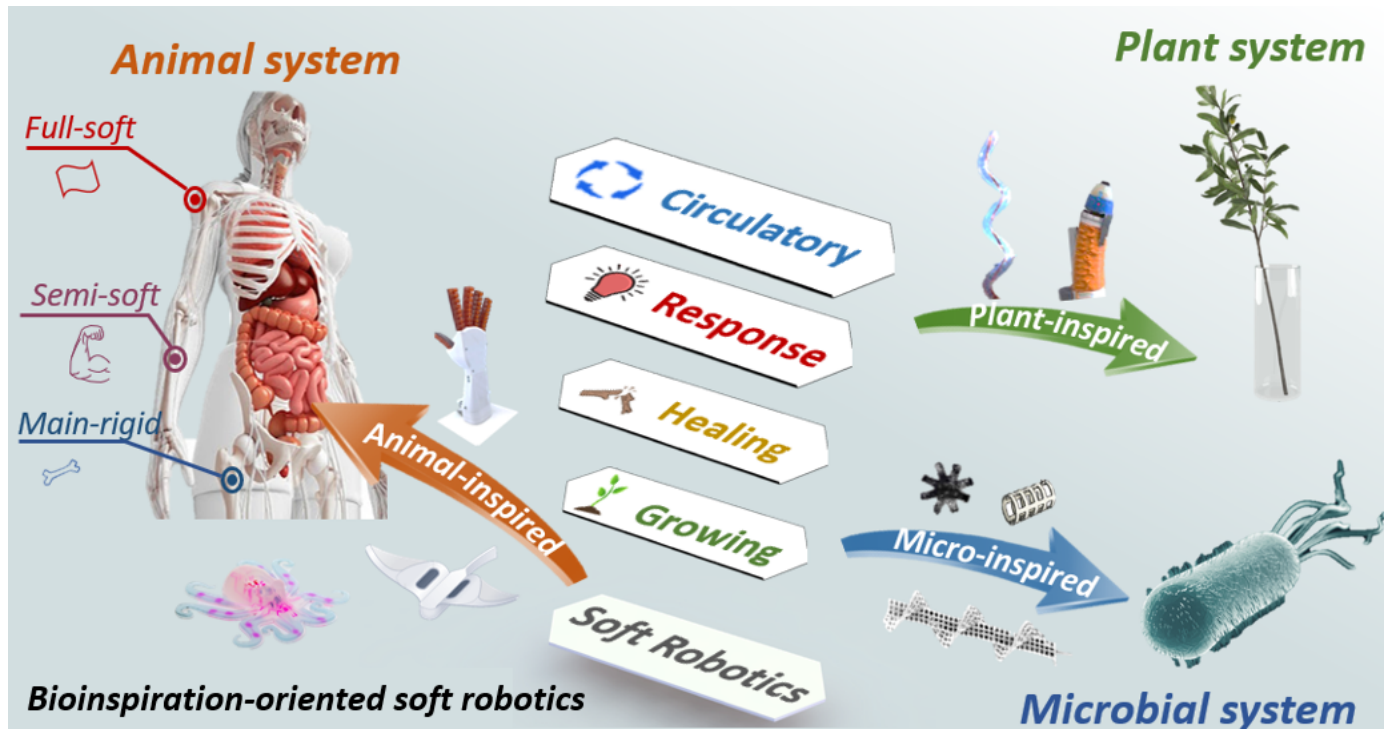

**Fig. 1.** An overview of this Review. A bio-function-based classification is proposed that contains the SG-SR, SH-SR, SR-SR, and SC-SR. These four categories mimic the macro functions of living creatures: movement, sensitivity, growth, reproduction, respiration, excretion, and nutrition.

Nature answers. A well-known knowledge is that all living creatures carry out similar life processes, including movement, sensitivity, growth, reproduction, respiration, excretion, and nutrition function [41] (Abbreviated as in Table II). With intelligent collaborations of these functions, creatures present the attractive biodiversity, which will be the final goal of bioinspired soft robots development. Learning from these functions, a macro-level soft robot classification can be summarized into four categories: self-growing soft robots (SG-SRs), self-healing soft robots (SH-SRs), self-responsive soft robots (SR-SRs), and self-circulatory soft robots (SC-SRs). SG-SRs refer to the growth function presenting the growing ability as creatures growing up. SR-SRs mimic the motion and sensitivity functions to sense environmental stimuli and respond via corresponding reactions. SC-SRs and SH-SRs simultaneously play a role in a metabolic system (containing reproduction, respiration, excretion and nutrition functions), performing creatures' circulatory function and self-healing ability, respectively (Fig. 1).

## III. WHAT CAN WE LEARN FROM CREATURES?

### A. Self-growing Soft Robots (SG-SR)

Self-growing ability, as a basic biological function of living organisms, represents processes of body-shape growth and body-function improvement. Although species grow up with various biological principles, the process can be microscopically regarded due to cell enlargement, cell division, and cell differentiation [42]. Learning from creatures, SG-SRs are defined as a class of soft robots that can build themselves to operate motion, attachment, exploration tasks, etc. SG-SRs aim to reproduce growth phenomena by self-building technologies to mimic the detailed and complicated cell activities behind them. For instance, shoot growth and root extension common in the plant kingdom (Fig. 2).

Generally, the growth ability is implemented by the incremental addition of new material or body length to the SG-SRs. Their fundamental design principle can be summarized as follows: Following different functions contributing to self-growth, structures of SG-SRs contain the stable base unit, elongation unit, and the orientation-adjustment unit, which corresponds to the controllability of the entire growth process, the elongation ability, and the guidance ability to control the growth direction, respectively. Regarding the bioinspired design principles, concepts that reproduce the above-mentioned abilities can be realized from various perspectives. For example, from an actuation-design perspective, a shoot growth-like SG-SR can

**TABLE I**

*Existing review articles on soft robots in the domain of different research perspectives.

| Areas | Design | | Fabrication | | Modeling | |
|---|---|---|---|---|---|---|
| *Perspective* | *Actuation methods* | *Application scenarios* | *Manufacture technologies* | *Material types* | *Dynamic analysis* | *Motion strategies* |
| *References* | Lee et al. [29] Zaidi et al. [30] Walker et al. [31] Hughes et al. [32] S. Kim et al. [170] | Poly et al. [33] Stokes et al. [34] Raizman et al. [35] Laschi et al. [171] | Cho et al. [36] Yang et al. [37] Joyee et al. [38] Li et al. [169] | Elango et al. [4] Li et al. [39] Giordano et al. [172] | Das et al. [40] | Calisti et al. [2] |
| *Categories* | ◆ Pneumatic ◆ Thermal ◆ Electric ◆ Optical ◆ PH ◆ Magnetic ◆ Chemical ◆ Wires | ◆ Gripper ◆ Crawling ◆ Rolling ◆ Jumping ◆ Swimming ◆ Arti. muscle ◆ Deliver ◆ Manipulator | ◆ FDM ◆ DIW ◆ SLS ◆ Inkjet ◆ SLA ◆ Origami ◆ Kirigami ◆ Molding | ◆ SMA ◆ SMP ◆ DE ◆ IPMC ◆ Hydrogel ◆ Silicon rub. ◆ Magnet ◆ Piezo/Tribo | ◆ Experiment ◆ FEA ◆ M. of matt. ◆ Large def. ◆ Empirical ◆ Para. study | ◆ Control ◆ Optimize ◆ Plan ◆ Coordi. ◆ AI |

*Abbreviations in Table I:
FDM:Fused Deposition Modeling, DIW: Direct Ink Write, SLS: Selective Laser Sintering, SLA: Stereolithography, SMA: Shape Memory Alloy, SMP: Shape Memory Polymer, DE: Dielectric Elastomer, IPMC: Ion-exchange Polymer-Metal Composites, Silicon rub.: Silicon Rubber, Piezo/Tribo: Piezoelectric material/ Triboelectric material, FEA: Finite Element Analysis, M. of matt.: Manufacture of materials, Large def.: Large deformation, Para. study: Parametric study, Coordi.: Coordinated motions, AI: Artificial Intelligence.

**Table II**

ABBREVIATIONS

| *Bio-function-based soft robot- basic level* | | CG | Circulatory self-growing |
|---|---|---|---|
| SG-SR | Self-growing soft robot | CH | Circulatory self-healing |
| SH-SR | Self-healing soft robot | *Bio-function-based soft robot- 3rd-order app. level* | |
| SR-SR | Self-responsive soft robot | RHC | Responsive self-healing circulatory |
| SC-SR | Self-circulatory soft robot | RGH | Responsive self-growing and healing |
| *Bio-function-based soft robot- 2nd-order app. level* | | CGH | Circulatory self-growing and healing |
| RH | Responsive self-healing | RGC | Responsive self-growing circulatory |
| RC | Responsive self-circulatory | | |

go across a long distance by self-growing without moving the whole body [43] (Fig. 2a). The base unit provides spaces for a gas system, generating air pressure for growth movements and a roll of stored tubes. The elongation and orientation -adjustment units refer to the stored tube and the control chamber. Pressurized by an air pump and bent by inflating chambers, the stored tubing and its moving directions can be lengthened and controlled. From the structure-design perspective, a vine-inspired SG-SR [44] is designed with a backbone structure, known as a typical structure to create shapes and movements for soft robots [45] (Fig. 2b). A spacer, backboned-springs design the base, elongation, and orientation

-adjustment units and tendons to grasp positions tightly, elongate its body, and adjust growing orientations, respectively. Directional extensions or compressions can be realized by releasing or pulling the three tendons, demonstrating a straightforward design of SG-SRs. From the manufacturing perspective (Fig. 2c), a root-like robot can build itself by depositing PLA materials [46] (i.e., 3D-print itself). The 3D printer stored PLA and depositing technology corresponds to the base, elongation, and orientation-adjustment unit. With this material addition method, soft robots can lively reproduce root-growing processes, realizing self-fabricate and self-exploration simultaneously. From the applicability perspective to output forces interacting with environments (e.g., overcoming resistances to take roots in soil), a skin-sliding principle-based SG-SR is designed to counter resistances from soils and particles [48] (Fig. 2d).

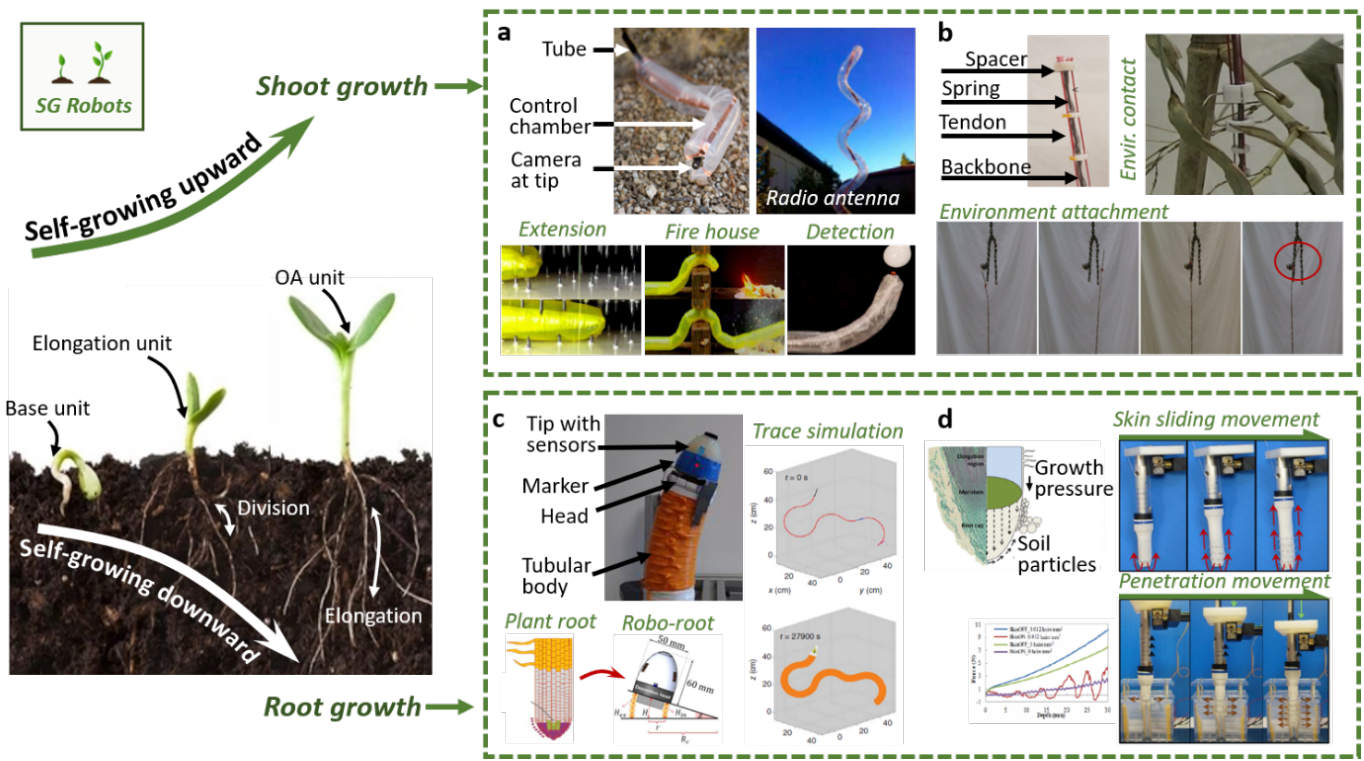

**Fig. 2.** Self-growing soft robots. a) A shoot growth-like SG-SR that can elongate and operate different tasks [43]. b) A vine-inspired SG-SR can automatically attach environmental objects [44]. c) A root-inspired SG-SR that can elongate itself by trace planning [46]. d) A skin-sliding principle-based SG-SR can grow into the soil by performing penetration movements [48].

The above-reviewed works realized the growth function from different perspectives. The shoot growth-like SG-SR (Fig. 2a) well illustrates the growth process, which has a maximum growth length of 72 m and a maximum growth speed of 10 m/s. Its pneumatic actuation and chamber designs contribute the impressive performances while enabling it to interact with environmental obstacles compliantly. However, when facing forcefully antagonizing the environment, the motor actuated SG-SR (Fig. 2d) performs unique advantages. Although the growth displacement (~40 mm) is limited by motor powers, it can break into the ground (soil with particles) with a speed of 1 mm/s and a force of 11.31 N. It demonstrates the satisfactory environmental interaction capability. Regarding the diversity of growth shapes, the 3D-printing actuated SG-SR (Fig. 2c) can potentially grow into any predetermined shape, and in its current stage, it has a maximum growth length of ~24 cm with a curvature radius of 17.5 cm. More quantitative self-growing ability comparisons can be found in Table III.

The unique characteristics of SG-SRs are: the elongation unit lengthens in the front while the base unit does not move when the soft material is added to the body at the other end [47]-[55]. This feature enables the robot to stick on the base stably (e.g., ground, soil). At the same time, the head of the robot can keep going forward by inserting materials. Another merit of SG-SRs is the deformability and compliance with the environment that enable them to go through narrow terrains (e.g., 10% of their widths [43]). Once the robot gets through the narrow gap, it will restore its size later on movement when the squeezing pressure is taken off from the environment. However, due to the essence of the growing phenomenon, SG-SRs are born with limited movability, resulting in constrained workspaces.

Their clear pros lead to targeted application scenarios such as field explorations, rescuers in complex environmental conditions, and even biomedical operations, whereas obvious cons cause limited abilities to react to environmental stimuli. Overviewing existing SG-SR studies, as mentioned before, roboticists make efforts mainly to realize intuitional growth behaviors. However, by developing more innovative methods (e.g., new smart materials with creative structure designs), SG-SRs with self-sensing and self-reacting abilities can be further realized to mimic the basic cell activities for growth.

### B. Self-healing Soft Robot (SH-SR)

The fascinating vitality of living creatures is illustrated by their tolerances and recoverability for damages taken from harsh nature. So should robots. This is known as the self-healing function that roboticists learn from creatures and endow SH-SRs with [64]-[68]. Existing review articles have defined and summarized SH-SRs from various perspectives, such as materials [56]-[59], actuators [60]-[61], and structures [62]-[63]. However, learning from creatures provides us with a new understanding of SH-SRs. According to different injury degrees and body regions, soft robots' self-healing can be summarized into two types that are the function-recovery and damage tolerance (Fig. 3). The function-recovery refers to the process that soft robots recover from function-affected injuries (e.g., Radius/ Ulna fractures.). Damage tolerance, also known

### Table III
#### COMPARISONS OF SG, SH, AND SR-SRS

| Literatures (selected) | Self-growing function | | | |
| --- | --- | --- | --- | --- |
| | Diameter (mm) | Growth length (m) | Growth speed (m/s) | Growth direction |
| Hawkes et al. [43] | 25 | 72 | 10 | 3D |
| Wooten et al. [44] | 16 | 1.33 | - | 3D |
| Dottore et al. [46] | 50 | 0.1-0.16 | 6.7x10⁻⁵ | 3D |
| Sadeghi et al. [48] | 20 | 0.04 | 10⁻³ | 1D |

| | Self-healing function | | | |
| --- | --- | --- | --- | --- |
| | Healing method | Healing time (hrs) | Healing efficiency | Healing condition |
| Terryn et al. [68] | Heating | 1-72 | 93.4-99 % | 80 °C |
| Wallin et al. [72] | Ultraviolet | 0.0083 | ~100 % | 9 mW·cm⁻² |
| Liu et al. [61] | Ultraviolet | 1 | 67 % | 1.5 W·cm⁻² |
| Cao et al. [73] | PH | 3 | 21/17.4 % | PH 1.3/13 |

| | Self-responsive function | | |
| --- | --- | --- | --- |
| | Stimuli type | Actuation time (s) | Velocity (bl/s) | Frequency (HZ) |
| Huang et al. [81] | Temp. | 0.22 | 0.1-1 | ~0.8-5 |
| Li et al. [93] | Electrics | 0.2 | 0.69 | ~0.1-14 |
| Huang et al. [97] | Magnet | 0.2 | 0.16-1 | 1-5 |
| Li et al. [95] | PH | 60-120 | 0.2 | - |
| Rogóź et al. [115] | Optics | 0.16 | 0.01-0.05 | ~6.25 |
| Yang et al. [120] | Comb. | 0.009 | 9-18 | ~1 |

as damage resilience [69], refers to the ability to prevent injuries from affecting their basic functions (e.g., the finger joint local injury does not affect the overall hand function).

Function-affect damages can be regarded as three types: the nerve-injury, muscle-injury, and connect-injury, which represent damages to robotic circuit system, actuation system, and structure connection, respectively. As the skin recovers from cuttings, a type of SH-artificial skin [73] can continue sensing touches, pressures, and strains after self-healing from circuit cuttings (Fig. 3a). This phenomenon is enabled by a fluorocarbon elastomer-fluorine-rich ionic liquid composition material that can recover its ionic conductivity through highly reversible ion-dipole interactions. It demonstrates the essence of neurological recoveries. Regarding the muscle-injury, a soft manipulator [78] can operate bending tasks after self-healing from various types of injuries, including local cuttings in different directions and global fractures (Fig. 3b). Moreover, existing McKibben muscles, graspers, and expansion actuators can also realize the muscle-injury recoveries by the material self-coating principle [68]. These SH-SRs lively reproduce the regeneration function of muscle fibers. In terms of the connect-injury, an SH soft structure [71] can reconnect itself after the global fracture (Fig. 3c), which reveals the structural reconnection phenomenon (e.g., animal bone-fracture treatment and plant grafting). It can be further applied in intelligent civil engineering scenarios, such as structure self-cure, structure-reinforcement, and structure self-connection.

The damage tolerance, regarded as an ability to tolerant regional and mild hurts from affecting basic robotic functions, is commonly found in creatures' self-protection activities, such as blood coagulation for wound curing. Methods that realize the damage tolerance can be summarized in two ways: wound-propagation prevention and rapid recovery [74]. Though pierced by a needle, a type of SH soft manipulator [70] can still operate bending motions (Fig. 3d). Moreover, regarding rapid-recovery, a type of SH soft muscle actuator [72] can leak resin and seal the wound by ultraviolet stimuli after getting a knife cut, which is like bleeding and blood coagulation (Fig. 3e).

Existing Reviews provided careful considerations of all evaluation factors of self-healing materials [69]. However, in this work, from the bio-inspiration perspective, the healing time and efficiency (defined as the ratio of Young's modulus after healing and before damage) are the important evaluation factors of the self-healing function. Table III shows common methods to realize the self-healing function: heating, ultraviolet (UV), and chemical. Heating is a well-studied method that can realize self-healing in a temperature range of 50-100 °C and a healing time of 1-72 hours [68]. Although the healing efficiency is satisfactory (more than 90%), the temperature requirements are challenging to reach in the natural environment. With this regard, the ultraviolet method can realize damage tolerance with a healing time of 30 s and impressive efficiency of ~100% [72]. It is more reasonable to let an SH-SR self-heal under the sunshine. It should be noted that, to heal the function-affected damage, the healing efficiency of the ultraviolet method is limited. Moreover, the pH healing method allows bio-inspired soft robots to heal themselves by making use of their living environments. SH-SRs can recover from seawater, pH 1.3 solution, and pH 13 solution with the healing efficiencies of 18.8 %, 20.9 %, and 17.4 %, respectively [73].

SH-SRs present prominent characteristics that can quickly recover from damages, demonstrating a robotics improvement---taking severe injuries does not mean failures anymore. Pros mainly lie in the unique material functions of SH-SRs. Several potential applications can be proposed: 1) Cell division SH-SRs. By the self-healing function, cell-inspired soft robots can recover their "cell membranes" after dividing into several sub-robots. 2) Disaster prevention and mitigation in civil engineering. SH soft structures can be further implemented as architectural components or sensors, which can cure themselves after taking damage from extreme scenarios (e.g., earthquakes, typhoons). Regarding the cons of the SH-SRs, curing efficiency can be further improved, "sequelae" after self-healings should be further investigated to ensure the applicability, and theories for cure predictions should be further developed.

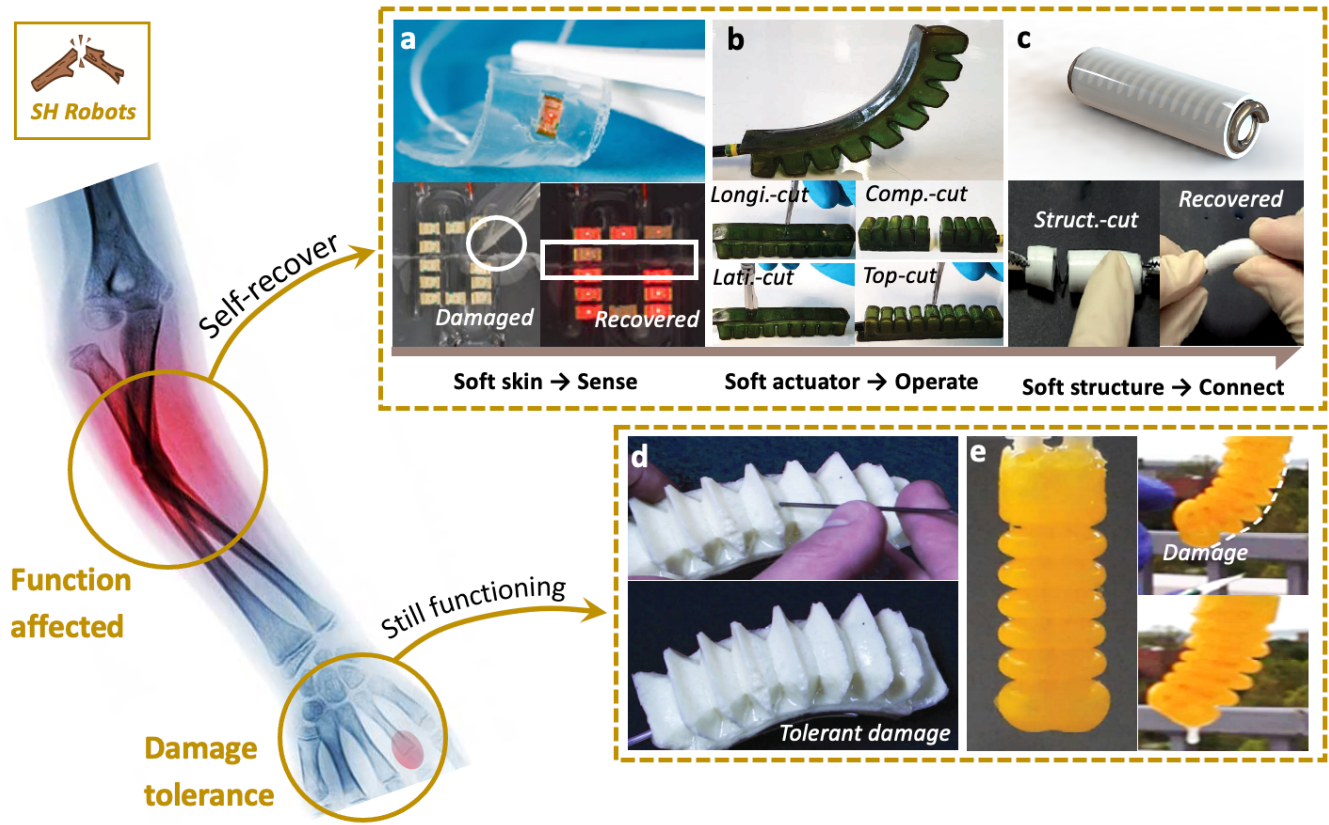

**Fig. 3.** Self-healing soft robots. a) A soft skin-inspired sensor that can work after recovering from damage [73]. b) An SH-soft actuator can normally operate tasks by bending motions after self-healing from different damages [68]. c) A shape memory alloy spring embedded structure sample can recover from fractures [71]. d) An SH-soft manipulator can tolerate damages during its operations [70]. e) An SH-synthetic antagonistic muscle actuator can tolerate damages during its operations [72].

## C. Self-responsive Soft Robot (SR-SR)

To sense, adapt and further interact to the natural environment, creatures are born with sense-response ability, such as the excitability of simple organisms (e.g., single-cell) and the reflex arc of complex organisms (e.g., nervous system) [75]. Humans are therefore inspired to design control systems for traditional robots that contain the sensing unit (e.g., sensors), control transmission unit (e.g., SCMs), and the response unit (e.g., motors) [76]. However, thanks to the unique characteristics of smart, flexible materials, soft robots make these units and their boundaries no more rigid. SR-SRs are bioinspired soft robots that can sense various environmental stimuli and perform corresponding motions. The significant difference between SR-SRs and other categories in this Review is that SR-SRs are targeted to actuate motions responding to the sensed stimuli. Based on the stimuli types, shape-memory materials, tribo/piezoelectric nanogenerator, dielectric elastomers, magnetic actuations, ion-exchange polymer-metal composites, and optic actuations are applied to design SR-SRs sensitive to the temperature, mechanical force, electricity, magnetic fields, pH, and optical stimuli [121]-[132].

Shape memory material [77]-[79] is known for its thermal inductivity. Designed by shape-memory materials, untethered soft jumpers, rollers, and walkers can perform motions at biological speeds (Fig. 4a) [81]. Despite the temperature awareness, shape-memory materials are also sensitive to light, electricity, magnet field, chemical signals, etc. [80]. Triboelectric nanogenerators [82]-[87] and piezoelectric nanogenerators [84] have been studied as green-energy sources and energy harvesting devices sensitive to mechanical forces such as frictions and deformations. Triboelectric nanogenerators have recently been applied in soft robots as friction-sensitive material generating e-signals that can detect surface roughness, motion processes, and deforming conditions of the soft robots [87] (Fig. 4b). Dielectric elastomers [88]-[93], known as a type of high-voltage awareness soft materials, are widely applied to design soft robots, such as the fast-moving soft fish (Fig. 4c), which can bear extremely high pressure under the deep sea (e.g., the Mariana Trench). The working principle can be briefly illustrated: negative and positive charges accumulate on both sides of the elastomer under a high voltage field, and the Maxwell stress can be generated to deform the dielectric elastomer membrane by performing bending motions. Regarding the magnetic fields sensitive SR-SRs, a soft fish designed with magnet-awareness materials can perform bioinspired swimming motions [97](Fig. 4d). The magnet-induced soft robots can be accurately controlled to perform deformations and further movements [98]-[106]. Note that, due to the accuracy and body-size-independency characteristics of the magnetic field, SR-SRs in cell-scale (known as micro-soft robots) are always realized by this method [107]-[114]. As a critical chemical factor, pHs can be sensed by ion-exchange polymer-metal composite materials deformed by environmental pH condition changings [94]-[96]. The ion-exchange polymer-metal composites-based soft robot is applied to deliver drugs through environments with different pH conditions (Fig. 4e). Additionally, the light-awareness soft materials with constant light stimuli can be triggered to realize degree-of-freedom control purposes, namely soft robot motion controls. A snake-inspired soft robot is designed with graphene-based light-driven materials, which is well controlled by accurately constraining degrees of freedom of the robot joints [115](Fig. 4f).

It should be noted that the above-reviewed actuation methods have been widely studied and reviewed from various perspectives [29]-[32]. Selected studies are compared to illustrate the bio-inspired SR function, which can be evaluated by the response-ability and kinematic performance(see details in Table III). The sensing-response activities can be divided into emergency-response, normal-response, and deferred-response. The combustion actuation method represents the emergency response, which provides a transiently high speed of 9-18 body lengths per second in about 9 ms [120]. Such rapid sensing-response phenomenon can be found in the hunting, escaping, knee jerk reflex, etc. The temperature, optics, magnet, and electric actuation methods contribute to the normal response simultaneously.

Although these methods respond to different stimuli, with a response time around 0.2 s and a generated velocity range of 0.01-1 bl/s, SR-SRs can sense external signals and react to them at a normal speed [81]-[115]. Methods for the normal response fulfill SR-SRs' fundamental motion requirements. Regarding the deferred response, the pH actuation method (respond to the pH changes in 60-120 s resulting in a velocity of 0.01-0.05 bl/s) can vividly mimic the slow response due to hormonal changes in creatures' bodies.

Regarding different responsive actuation methods, the pros and cons are varied, such as the shape-memory materials can endure high temperature and therefore be deformed, whereas the dielectric elastomer may lose its functions under this condition [29]-[32]. Magnets-responsive actuation can be accurately controlled with relatively low output forces, whereas the combustion actuation method can output extremely high forces with difficulty to control [117]-[120]. However, from the bioinspiration perspective, SR-SRs can be modeled by deformation theory and stimuli corresponded mechanics. SR-SRs are sensitive to stimuli, which vary in different environments leading to control-instability.

SR-SRs, the widest studied type of soft robotics, have been widely proven to be applied in various scenarios, such as the bio-fish that can explore great seas [93], the bio-worm that can shuttle back and forth in the soil [167], the bio-bird that can fly to the sky [168], and the micro-robot that can operate tasks inside the human body [95]. However, due to the limited robustness and environmental sensitiveness, SR-SRs face challenges in achieving more practical and reliable applications to compete with or complement rigid robots.

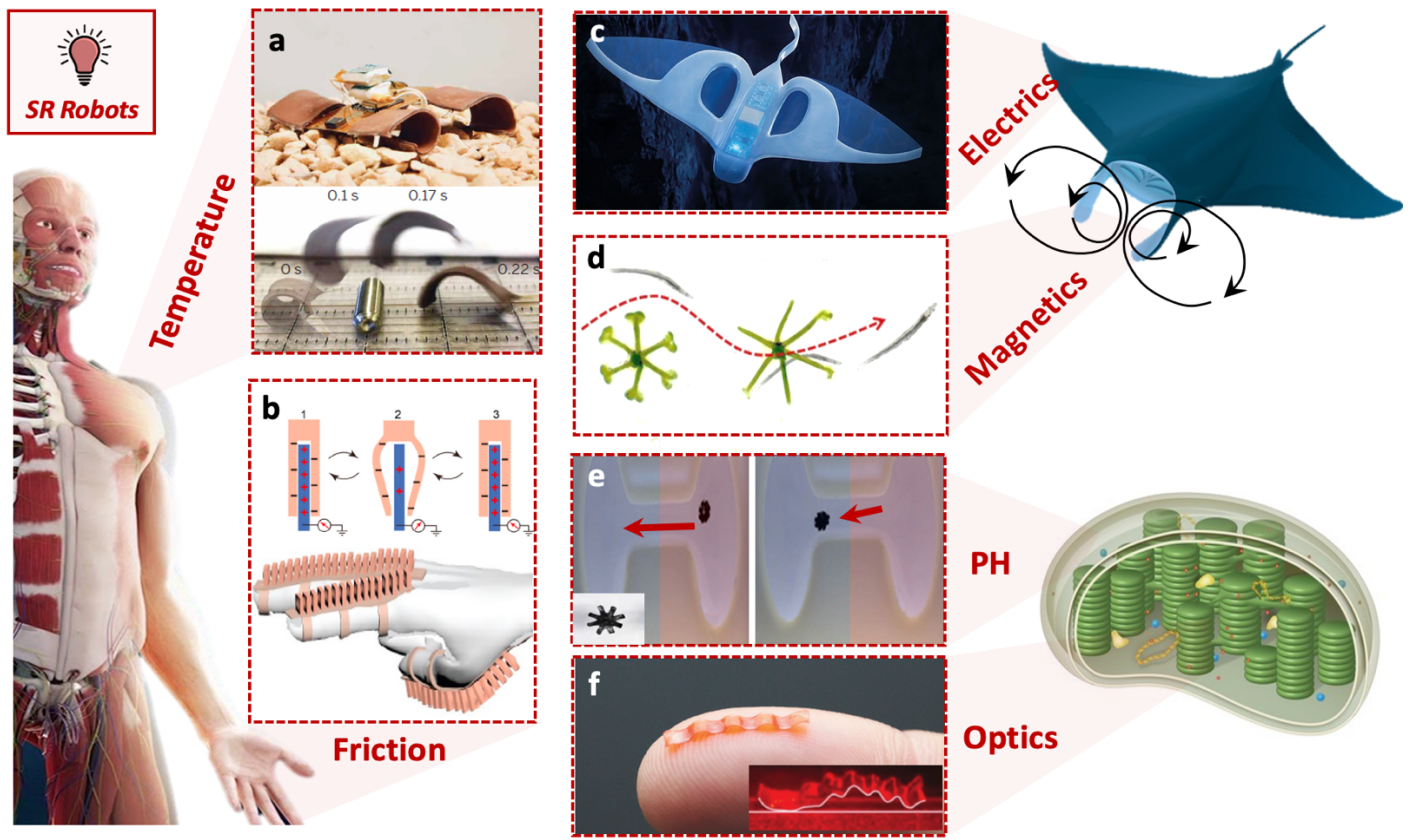

**Fig. 4.** Somatosensory self-responsive soft robots. SR-SRs can respond to various kinds of environmental stimuli, such as the human body that can respond to a) temperature- Shape memory material-based soft robots: a soft walker, a soft jumper, and a soft roller [81], b) mechanical forces- Triboelectric nanogenerator enabled soft gripper [87], c) electricity- Electronic fast-moving fish driven by dielectric elastomers [93], d) magnetic fields- Worm-inspired soft robot driven by magnetic fields [97], e) pH- ion-exchange polymer-metal composites enabled drug-delivery soft robot driven by environmental pH changes [95], and f) optical stimuli- Snake-inspired soft robot driven by light stimuli [115].

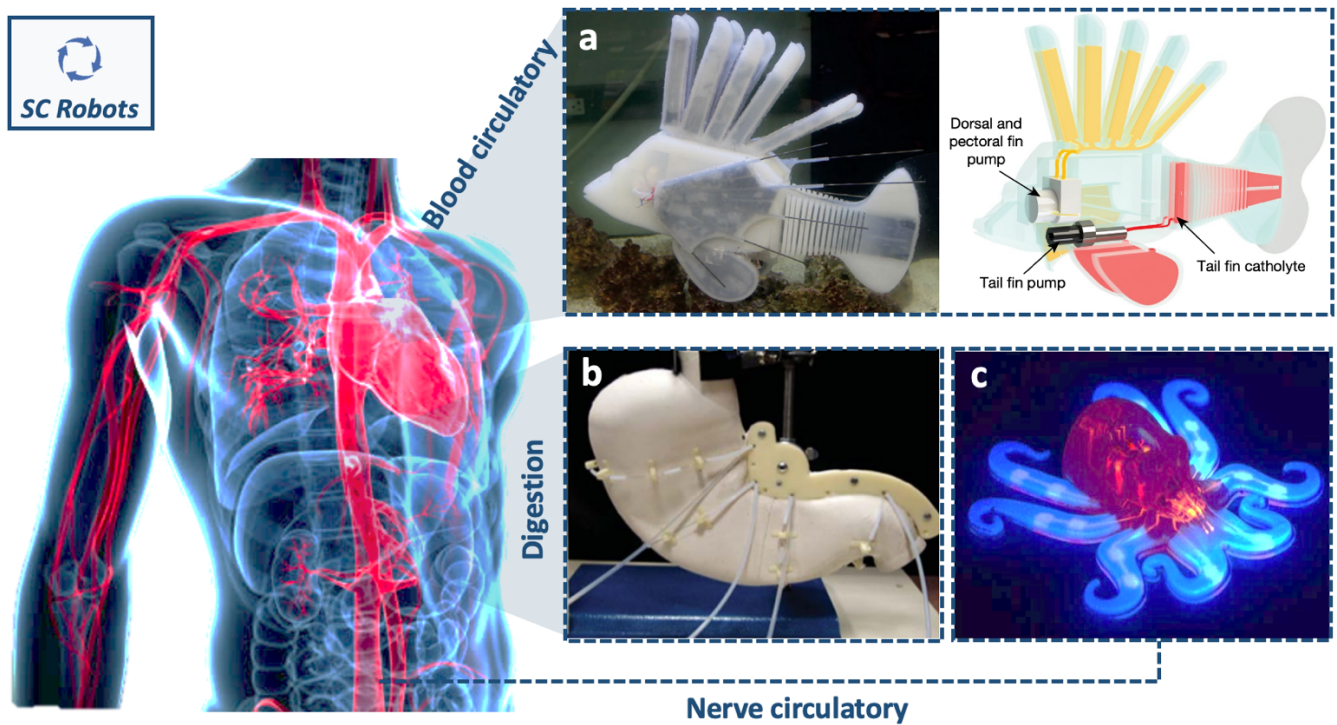

**Fig. 5**. Self-circulatory soft robots. a) An electrolytic vascular system for a soft fish and a schematic diagram of its working principle [134]. b) An artificial stomach that can mimic digestion processes [135]. c) A full-soft octopus-inspired robot designed with fuel channels [136].

### D. Self-circulatory Soft Robot (SC-SR)

The self-circulatory function mimics three main functional processes of creatures: sensitivity function, respiration, excretion, and nutrition function, which are contributed by the digestion, circulatory, breath, and nerve systems. SC-SRs are born with these functions implementing the comprehensiveness of the soft robotics area.

The blood circulatory, a basic life process of advanced animals playing important roles in nutrition transportation, hormonal regulation, and energy delivery, is reproduced by an electrolytic vascular system for a soft fish (Fig. 5a). The vascular system integrates hydrodynamic transmission, actuation, and energy storage to reproduce blood circulatory, motion ability, and nutrition function. Learning from creatures facilitates the development of this kind of subversive invention. Not limited to the closed-loop, soft pumps (or systems) that can trigger open-looped circulatory phenomenon (e.g., excretion, urinary) are also investigated [133]. An artificial stomach can mimic peristaltic movement to provide testing spaces for endolumininal robotics [135] (Fig. 5b). Similar functions can also be realized by combustions [117]-[120]. For instance, a combustion-enabled soft pump can generate powerful output pressures, providing impressive potential circulatory applications [119]. As another significant open-loop circulatory model, the nerve system is realized by a fuel-channel enabled full-soft octopus-inspired soft robot. Following a similar nerve-signal transmission principle, the fuel can be delivered to reaction sites following the printed fluidic networks [137] (Fig. 5c). Moreover, as blood is circulating in vivo, a fully integrated fluidic circuitry is proposed to realize SC that can be efficiently controlled [116].

SC-SRs have functional and systematic advantages at the preliminary stage, with limited cases and manufacturing technologies. Namely, SC-SRs face challenges in designing diverse circulatory systems, finding efficient power sources for circulations, and extending their functions to various application scenarios. The functional advantage refers to their potential to be carriers that can combine various actuation methods. Specifically, the integrated systematic structures endow their possibility to be a full-soft robot, a deep-bioinspiration goal of soft robotics developments.

Creatures have evolved with optimized, experienced, and efficient functions and systems, which roboticists can learn to design soft robots. Growing ability, recoverability, response-ability, and circulatory-ability contribute simultaneously to the bio-functional integrity of soft robotics. These abilities can be combined to design various deep-bioinspired soft robots with satisfactory environmental adaptabilities and practical applicabilities. To illustrate ***what we learn from creatures***, following bio-functions: motion, sensitivity, growth, reproduction, respiration, excretion, and nutrition functions, we classified soft robots into four categories that are SG-SRs, SH-SRs, SR-SRs, and SC-SRs.

## IV. How do we learn from creatures?

Like creature evolutions, soft robotics developments can possess uniformity and diversity. Regarding uniformity, a soft robot can be considered one (or a combination) of the reported four categories as they are derived from basic bio-functions. Regarding the diversity, creative methods realizing these categories are continuously emerging at an impressive speed, such as new materials responding to environmental stimuli, new structural designs providing operation supports, and new bioinspired movements fulfilling various engineering requirements [153]-[162]. From the bioinspiration perspective, further insight into soft robot development can be obtained, illustrating how we learn from creatures.

The bioinspiration classification defines different soft robot types and boundaries between every category. However, a well-known fact is that robots can possess multifunctional properties, as intersections of two or more categories. For example, in terms of the actuation classification, advanced artificial muscles are actuated by electricity and hydraulic pressures simultaneously [63]. In application-oriented classification, drug delivery micro-robots have deliverability and swim ability simultaneously in the liquid medium [95]. So is the proposed bioinspiration classification method: the intersections of bio-functions that lead to biodiversity. Taking advantage of the bioinspiration-oriented classification, the intersections illustrate the current level of soft robot development while providing potential developing trends (Fig. 6). Combining four basic categories (SG, SH, SR, and SC) generates six types of 2nd-order applications, four types of 3rd-order applications, and a 4th-order application, which can be regarded as the soft robotics progress with increasing intersection-order. Existing literature and potential applications are marked by solid-line and dashed-line frames, respectively.

Here, we analyze and discuss these intersections. Increasing the order number indicates the remarkable progress of soft robotics development. From the bioinspiration perspective, fundamental soft robotics investigations are in the 1st-order level, which mainly focuses on realizing basic bio-functions by developing creative transmission supporters (e.g., origami robots [138]-[139], kirigami robots [140]-[141]), creating intelligent soft materials responding to various stimuli whilst investigating corresponding control theories (e.g., dielectric elastomer large deformation [142]-[147]), and imitating presentational bio-motions (e.g., swimming, jumping, crawling, and grasping [148]-[152]). The 1st-order application level has been reviewed in the previous section of this Review.

Regarding the 2nd-order level, soft robots have coupled bioinspiration functions integrating two basic functions. Six intersections yield the responsive self-growing (RG), self-growing and healing (GH), responsive self-healing (RH), responsive self-circulatory (RC), circulatory self-growing (CG), and circulatory self-healing (CH) soft robot, respectively. Photosynthesis, as one of the significant botanical local functions, inspires the responsive self-growing soft robots (Fig. 6a). Not limited to response to optical stimuli, soft robots with RG can grow in response to various environmental stimuli (e.g., pneumatics [47]). Performing gentle and directional motions, RG soft robots can be applied as field explorers, gastrointestinal tract inspections, and even robot cells that can grow by stimuli. The GH soft robots refer to the growth process that needs growing and recovering from harsh nature. A novel self-replication and reconfigurable soft robot are proposed to demonstrate the cell-replication ability of GH soft robots [163] (Fig. 6b). The self-growing process realizes the body-enlarge process, while the self-healing process realizes the annexation process, membrane repairing, and membrane merging. GH demonstrates a significance, indicating that soft robots can struggle against environmental challenges and grow robust. RH soft robots are a type of well-studied 2nd-order application. Responding to environmental stimuli and therefore performing motions with the self-healing ability, existing studies have already reported optics-based, temperature-based, and electricity-based designs [61] (Fig. 6c). Unlike GH, RH refers to movability (how to adapt, move, and interact with the environment) with robustness instead of focusing on self-developments. RC contains the responding and self-circulatory abilities that refer to the motion and control systems.

It should be noted that the control system, which is different from classic control concepts, indicates the comprehensive nerve or fluidic system in robot bodies to control themselves. For instance, artificial muscles with nerve arcs can react to acupunctures performing contractions. Limbs generate motions with energy supplies with signals delivered through blood and nerve systems [136] (Fig. 6d). The above discussed 2nd-order applications are involved in existing literature, and due to the limited study of SC-SRs, CG and CH have not been further investigated. However, efforts should be taken to perfect the 2nd-order application studies. CG indicates the processes of circulatory system self-building, which can be regarded as a new manufacturing technology enabling circulatory systems to expand and distribute to required regions (e.g., nerve endings always distribute densely in important body regions). Therefore, CH corresponds to the recoveries of damaged nerve or blood systems, merging circulatory units to enlarge or reinforce themselves (Figs. 6e-f).

In the recent five years, the publication quantity of the 1st-order level soft robots is more than 200 per year [173], which SR-SR accounts for the majority. The motion-ability and sense ability the SR-SRs represent are two important functions that robotics researchers can most directly think of and endow robots with. With the continuous improvement of the basic bio-functions of soft robots, the four categories of the 1st-order level are well studied. Therefore, various 2nd-order applications are being investigated. Namely, the current stage of soft robotics development is stepping into the 2nd-order application level. To integrate two basic bio-functions achieving the 2nd-order application, properties of different 1st-order categories should be clarified. The SG, SH, SR, and SC functions represent the shape change capability, adaptability, maneuverability, and systematicness, respectively. RH can be realized by developing self-healing materials that can simultaneously respond to external stimuli to move (e.g., a self-healing light-driven shape memory polymer [61]). Existing SG-SRs are actuated by motors, which can evolve to RG once the actuators are replaced by self-response systems that can sense external stimuli [47]. Unlike other 2nd-categories, GH is more dependent on developing smart materials that can swallow environmental substances to build and heal themselves [163]. Replacing rigid circuit devices with more flexible self-circulatory systems can evolve SG, SH, and SR into GC, CH, and RC, respectively. However, it is challenging to develop self-circulatory systems. Potential solutions are building energy storage systems [134] and manufacturing microscale flow channels by the embedded 3D printing technique [136].

Regarding the 3rd-order application level, as a stage current studies have not reached yet (to the best of our knowledge), soft robots are systematic (global) bio-functions that demonstrate more environment-adaptable, intelligent and multifunctional bio-behaviors. There are four intersections for the responsive self-healing circulatory (RHC), responsive self-growing and healing (RGH), circulatory self-growing and healing (CGH), and responsive self-growing circulatory (RGC) soft robots, respectively. We outlook potential applications of every intersection to illustrate fascinating and unique applications of soft robotics, which rigid robots are hard to realize. RHC represents circulatory systems' self-recovery ability to respond to environmental stimuli. The body-shaping, by which bodybuilders train their muscles to enhance strength and health, can be regarded as a process of muscle fibers' fracture-recovery loop [164]. Therefore, muscles can be reinforced to adapt to higher loads for different purposes, such as intensive physical jobs. It reveals an innovative and intelligent artificial muscle application (Fig. 6g). The RHC muscle could be of different load ability by repeating fiber fracture-recovery training. This advanced concept of flexibility illustrates a creative interaction method to environmental stimuli and a self-learning process understanding *how strong they should be*. Unlike RHC, RGC focuses on the circulatory establishments under environmental stimuli. The conditioned reflex establishment is assumed as a neural connection formation process in responses to external stimuli [165], corresponds to one of the RGC applications (Fig. 6j). An RGC system can self-build its connections by stimulation training to learn *how to react to new stimuli.* Therefore, GHC can be regarded as a self-growing and self-healing process of the circulatory system (Fig. 6i). With GHC, soft robots can sense, respond to, and learn from nature to answer the question—*how to survive in the environment by themselves.* Moreover, grafting, a well-known agronomy technology, combines different plant varieties into a multi-properties type [166] like the RHC (Fig. 6h). Various materials actuating soft robots could be grafted together by the boundary-coating (SH process) response to specific external stimuli, which can be regarded as a manufacturing technology. With this RHC application, multifunctional soft robots can be efficiently designed and fabricated, illustrating *how to obtain new actuation methods.*

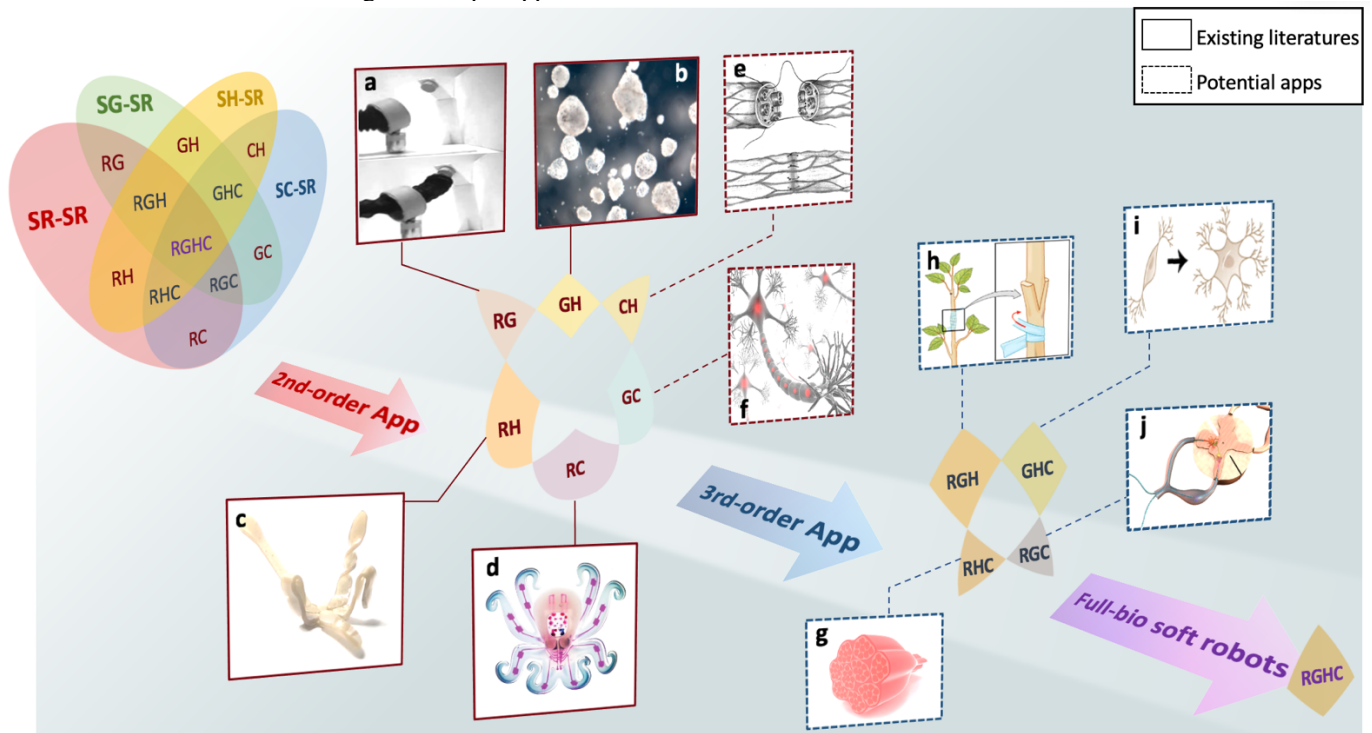

**Fig. 6.** Bioinspiration soft robots: overview and outlook. Category intersections lead to 4-level applications, which are the 1st-order application level including the SG, SH, SR, and SC, 2nd-order application level including the RG, RH, GH, GC, RC, CH, and GC, 3rd-order application level including the RHC, RGH, GHC, and RGC, and 4th-order application level that is RGHC. Existing or potential applications are illustrated for every intersected category, respectively. a) Responsive-self growing soft robot. A pneumatic responsive self-growing soft manipulator can reach targets through narrow terrains [47]. b) Self-healing-growing soft robot. A soft robot can reproduce and reshape its shapes by programming [163]. c) Responsive-self healing soft robot. A flower-inspired soft robot can bloom under optical stimuli and heal itself after getting damaged [61]. d) Responsive self-circulatory soft robot. The reported Octobot is actuated by fuel stimuli stored in its circulatory channels [136]. e) Circulatory self-healing soft robot. Recoverable channels or vessels that connect soft robots' components. f) Circulatory self-growing soft robot. Self-establishable nerve cells can automatically create or reinforce the nerve arc. g) Responsive self-healing circulatory soft robot. Artificial muscle fibers can train themselves by fracture-recovery loops. h) Responsive self-growing and healing robot. Under the photosynthesis phenomenon, different soft robot components can be connected, which is like the plant-grafting process. i) Circulatory self-growing and healing robot. Connections (e.g., flexible connections, rigid-soft connections, and inner channels) can be established, distributed, or reinforced automatically under continuous training from external stimuli. j) Responsive self-growing circulatory soft robot. Artificial nerve systems can realize nerve arc self-building, sensing and reaction to outside stimuli.

Instead of leading to confusion, these intersections take advantage of the function-based classification and provide bioinspiration-understandings of an all-along topic—"what will soft robots be like in the future?". Soft robots of systematic functions (3rd-order level) represent the applicability and intelligence improvements. Further, RGHC soft robots (4th-order level) can get environmental nutrients to grow, obtain experiences to adapt to nature, train themselves to get strong and learn to deal with various external stimuli.

## V. Conclusion

In this Review, we propose a bio-functions-based classification that contains SG-SRs, SH-SRs, SR-SRs, and SC-SRs covering basic bio-functions of motion, sensitivity, growth, reproduction, respiration, excretion, and nutrition functions, demonstrating **why** we learn from creatures. State-of-art technologies, characteristics, pros, cons, challenges, and potential applications of every category summarize **what** we learn from creatures. To find out **how** we learn from creatures, we extend beyond every category and further obtain the 2nd-order (including the RG, GH, RH, RC, CG, CH), 3rd-order (including the RHC, RGH, CGH, RGC), and 4th-order (RGHC) levels as inferences. Following these intersections, potential soft robots can be inspired, indicating futural development trends.

Different from other Reviews, this Review aims to macroscopically understand how bioinspired soft robots can be classified and what is the current stage and the developing trend of bioinspired soft robots. It should be noted that this work and other reviews complement each other. In this work, we cited limited literature, including Reviews, to help us explain the definition of each category (SG, SH, SR, and SC). Other reviews are valuable references to understand current technologies in soft robot studies of a specific category.

This bio-function-based soft robots classification provides a systematic understanding of current developments and envisions futural research.

## Acknowledgment

H. R. acknowledges the National Key R&D Program of China under Grant 2018YFB1307700 (with subprogram 2018YFB1307703) from the Ministry of Science and Technology (MOST) of China, the Shun Hing Institute of Advanced Engineering (SHIAE project #BME-p1-21, 8115064) at the Chinese University of Hong Kong (CUHK), the Key Project 2021B1515120035 of the Regional Joint Fund Project of the Basic and Applied Research Fund of Guangdong Province, and Singapore Academic Research Fund under Grant R397000353114.

Y. Y. acknowledges the China Scholarship Council scholarships (202006320349) at the Ministry of Education (MOE), China.

Z. H. acknowledges the National Key Research and Development Program of China (2021YFE0206200), and the Key Research and Development Project of Hainan Province (ZDYF2022SHFZ045).

P. J. acknowledges the Key Research and Development Plan of Zhejiang, China (2021C03181).

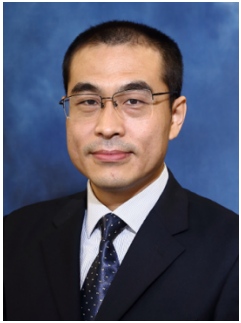

**Hongliang Ren** received his Ph.D. in Electronic Engineering (Specialized in Biomedical Engineering) from The Chinese University of Hong Kong (CUHK) in 2008. He has been navigating his academic journey through Chinese University of Hong Kong, UC Berkeley, Johns Hopkins University, Children's Hospital Boston, Harvard Medical School, Children's National Medical Center, United States, and National University of Singapore. Dr. Ren is a senior member of IEEE.

Dr. Ren's research interests include Biorobotics & intelligent systems, medical mechatronics, continuum, and soft flexible robots and sensors, multisensory perception, learning and control in image-guided procedures, deployable motion generation.

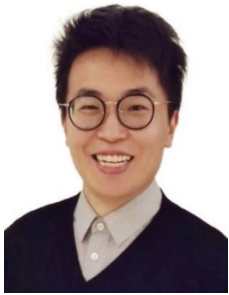

**Yang Yang** received his Ph.D. in Ocean Engineering from Zhejiang University, Zhejiang, China, in 2022. He is currently working as a Postdoctoral Research Fellow at the Chinese University of Hong Kong.

His research interests include the transient driving method (TDM) of the soft robotics, biomimetic soft robotics, the hydrodynamics of underwater soft robots, biomedical soft robotics, and large deformation mechanics of the flexible materials.

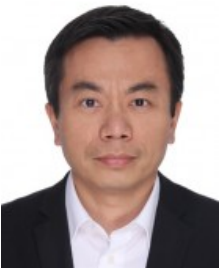

**Zhiguo He** earned his PhD degree from the University of Mississippi in 2007. Dr. He is a professor at ocean college of Zhejiang University, also serves as chair of department of ocean engineering.

Dr. He's research interests include the computational fluid dynamics (CFD), fluid dynamics of underwater robots, numerical modeling.

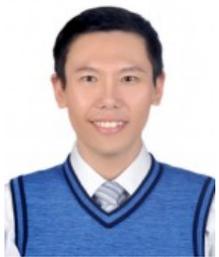

**Pengcheng Jiao** earned his MSc and PhD degrees in Civil Engineering from the West Virginia University and the Michigan State University in 2012 and 2017, respectively. Prior to joining the Zhejiang University as a Research Professor in 2018, Dr. Jiao was working as a Postdoctoral Research Fellow at the University of Pennsylvania.

Dr. Jiao's research interests include marine mechanical metamaterials, structural health monitoring (SHM) in ocean engineering, marine soft robotics, and artificial intelligence (AI) in ocean engineering.